\documentclass[10pt,twocolumn,letterpaper]{article}

\usepackage[pagenumbers]{cvpr} 
\makeatletter
\@namedef{ver@everyshi.sty}{}
\makeatother
\usepackage{tikz}

\usepackage{graphicx}
\usepackage{amsmath}
\usepackage{amssymb}
\usepackage{booktabs}
\usepackage{placeins}
\usepackage[normalem]{ulem}
\usepackage{multirow}
\usepackage{pifont}
\usepackage{subcaption,siunitx,booktabs}
\usepackage[font={small}]{caption}
\usepackage[misc]{ifsym}

\usepackage[pagebackref,breaklinks,colorlinks]{hyperref}

\usepackage[capitalize]{cleveref}
\crefname{section}{Section}{Secs.}
\Crefname{section}{Section}{Sections}
\Crefname{table}{Table}{Tables}
\crefname{table}{Tab.}{Tabs.}

\title{PaintHuman: Towards High-fidelity Text-to-3D Human Texturing \\via Denoised Score Distillation}
\author{
Jianhui Yu$^{1}$ \quad Hao Zhu$^{2}$ \quad Liming Jiang$^3$ \quad Chen Change Loy$^3$ \quad Weidong Cai$^1$ \quad Wayne Wu$^{2}$\\
{\small $^1$University of Sydney \quad $^2$Shanghai AI Laboratory \quad $^3$S-Lab, Nanyang Technological University }\\
{\tt\small 
jianhui.yu@sydney.edu.au \quad 
haozhu96@gmail.com  \quad 
\{liming002,ccloy\}@ntu.edu.sg
} \\
\vspace{-1mm}
{\tt\small 
tom.cai@sydney.edu.au \quad 
wuwenyan0503@gmail.com}
}

\begin{document}
\twocolumn[{%
\renewcommand\twocolumn[1][]{#1}%
\maketitle
\begin{center}
\vspace{-5mm}
\centering
\includegraphics[width=\linewidth]{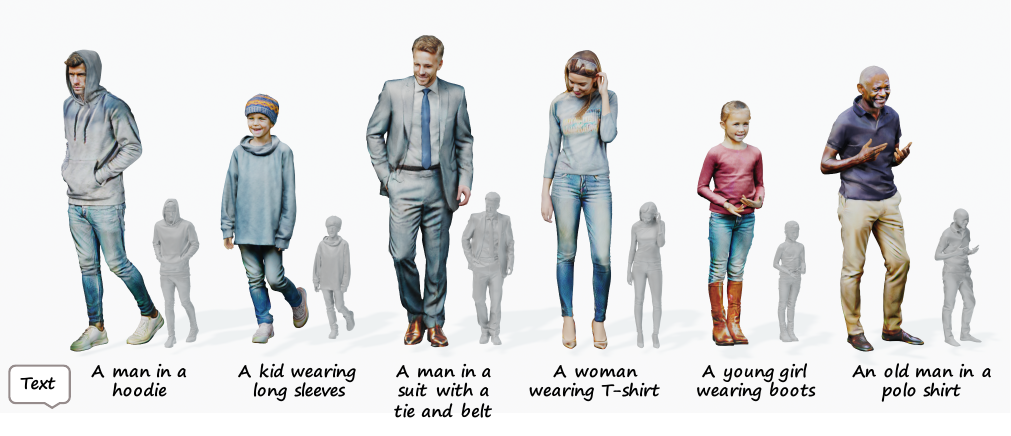}
\vspace{-10mm}
\captionof{figure}{Generated results of PaintHuman. Given textureless human meshes and textual descriptions as input, our model can generate high-quality and detailed textures that aligned to input geometry and texts.}
\end{center}%
}]

\begin{abstract}

Recent advances in zero-shot text-to-3D human generation, which employ the human model prior (\eg, SMPL) or Score Distillation Sampling (SDS) with pre-trained text-to-image diffusion models, have been groundbreaking. 
However, SDS may provide inaccurate gradient directions under the weak diffusion guidance, as it tends to produce over-smoothed results and generate body textures that are inconsistent with the detailed mesh geometry.
Therefore, directly leverage existing strategies for high-fidelity text-to-3D human texturing is challenging. 
In this work, we propose a model called \textit{PaintHuman} to addresses the challenges from two aspects. 
We first propose a novel score function, \textit{Denoised Score Distillation (DSD)}, which directly modifies the SDS by introducing negative gradient components to iteratively correct the gradient direction and generate high-quality textures.
In addition, we use the depth map as a geometric guidance to ensure the texture is semantically aligned to human mesh surfaces. 
To guarantee the quality of rendered results, we employ geometry-aware networks to predict surface materials and render realistic human textures.
Extensive experiments, benchmarked against state-of-the-art methods, validate the efficacy of our approach.

\end{abstract}

\section{Introduction}




Significant progress has been made in text-to-3D content generation. Some methods are proposed for general objects~\cite{dreamfusion, magic3d}, and some are specifically for 3D human avatars~\cite{dreamavatar, avatarclip, avatarcraft}.
3D human avatars are increasingly important in various applications, including games, films, and metaverse.
In this work, we focus on texturing a predefined human mesh with text prompts.

%
Success of recent methods rely on CLIP model~\cite{avatarclip} or text-to-image generation, which leverages the diffusion model~\cite{ddpm, sd}, and Score Distillation Sampling (SDS)~\cite{dreamfusion} combined with differentiable 3D representations~\cite{nerf, mipnerf}.
However, directly leverage existing strategies for detailed human avatar texturing in a zero-shot manner is challenging due to two reasons.
First, we find that SDS is a general-purpose optimization, which guides the loss gradient to a direction due to its weak supervision and unable to well handle unclear signal from the diffusion model. This issue results in generated human textures of low quality, including over-smoothed body parts and blurry garment details. 
Second, textures guided by text-to-image models are usually not semantically unaligned to either input texts or human mesh surfaces, resulting in missing textures or unaligned texture mapping for the geometry.



Recent works~\cite{avatarclip, avatarcraft} for human avatar texturing entangles the shape and texture generation, which leverage human-specific priors~\cite{smpl} for human body texturing.
To ensure the generated textures aligned to the given geometry, TEXTure~\cite{TEXTure} and Text2Tex~\cite{Text2Tex} utilize a depth-aware diffusion model~\cite{sd} to directly inpaint and update textures from different viewpoints, which could cause inconsistency when the input mesh has complex geometry.
Other methods such as Latent-Paint~\cite{latentnerf} or Fantasia3D~\cite{fantasia3d} apply SDS to update the loss gradient for consistent texture generation.
However, the issue associated with SDS has not yet been solved, \ie, the synthesized textures are non-detailed and over-smoothed, failing to be semantically aligned to the input texts.

Therefore, we propose \textit{PaintHuman} to address a primary issue associated with SDS. Our main idea is to denoise the unclear gradient direction provided by SDS loss. 
We handle this from two aspects. 
Firstly, we propose \textit{Denoising Score Distillation} (DSD), which introduces a negative gradient component to directly modify the SDS, which could correct the gradient direction iteratively for detailed and high-quality texture generation.
Then, to enable geometry-aware texture generation, we utilize geometric guidance which provides rich details of the mesh surface to guide the DSD precisely, and use spatially-aware texture shading models~\cite{realshading} to guarantee the quality of rendered visual results.

%
%
Specifically, DSD utilizes an additional negative pair of image and text.
The key idea is that by utilizing a negative image, \ie, an image with noise rendered from the last training iteration, we could reinforce the learning of the complex surface geometry to produce clear boundaries between different garments.
Besides, with the help of negative text prompts, the synthesized textures could be more semantically aligned to the input text.
Overall, the negative pair contributes a negative part to SDS, which controls the gradient direction by a weighted subtraction of the two input pairs, producing an effective gradient to address over-smoothed texture generation.
%
%
To further ensure textures semantically aligned to the complex avatar surface, we first use the depth map as guidance during the diffusion process for texturing, which provides the fine-grained surface details.
%
In addition, we follow~\cite{nvidia_extracting} to apply the Spatially-Varying Bidirectional Reflectance Distribution Function (SV-BRDF)~\cite{realshading} and coordinate-based networks~\cite{ngp} for geometry-aware material prediction.
With the help of differentiable rendering~\cite{rendering}, we could update the rendered human avatar and synthesized textures in an end-to-end fashion.

The contributions of our work are summarized as follows:
\begin{itemize}
    \item We introduce Denoising Score Distillation (DSD), a diffusion-based denoising score using negative image-text pairs for high-fidelity texture generation aligned to textual descriptions.
    \item We employ semantically aligned 2D depth signals and spatially-aware rendering functions for geometry-aware texture generation and realistic avatar rendering.
    \item Through comprehensive experiments, we prove the efficacy of our method over existing texture generation techniques.
\end{itemize}

\begin{figure*}[ht]
    \centering
    \includegraphics[width=0.9\linewidth]{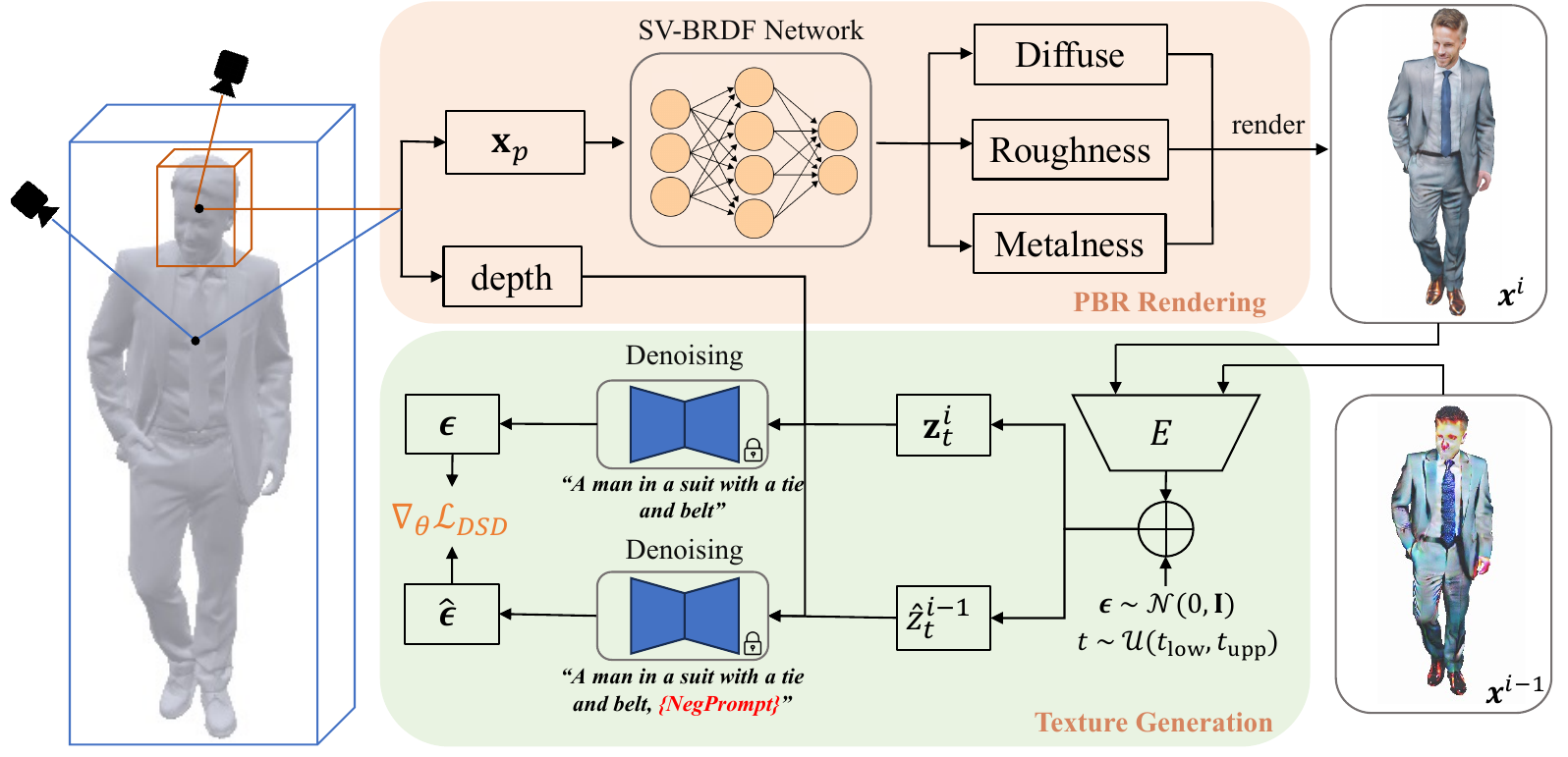}
    \caption{
    Overview of our proposed model. Our goal is to texture the human mesh given an input text and a mesh model. To achieve this, we propose Denoised Score Distillation with a negative pair of image and text prompts to guide the gradient direction for detail texture generation that is semantically aligned to the input text. We introduce depth signals to the diffusion process for complex garment texturing, and 
    a learnable network to estimate SV-BRDFs for albedo and material parameters learning. Finally, camera position is adjusted for refined detailing of the face region.
    } \label{fig:painthuman_pipeline}
    \vspace{-3mm}
\end{figure*}

\section{Related Work}
\subsection{Diffusion Models}
With the development of denoising score-matching generative models~\cite{sohl2015deep}, diffusion models have presented great success in a variety of domains such as image editing, text-to-image synthesis, text-to-video synthesis, and text-to-3D synthesis.
In the field of text-to-image synthesis, diffusion models have demonstrated impressive performance, especially the Stable Diffusion model~\cite{sd}, which is trained on a large number of paired text-image data samples with CLIP~\cite{clip} to encode the text prompts and VQ-VAE~\cite{vqvae} to encode images into latent space.
In our work, we use a pre-trained Stable Diffusion model to incorporate intrinsic image prior to guide the training of our texture generation network.

\subsection{3D Shape and Texture Generation}
There has been a recent surge of interest in the field of generating 3D shapes and textures. One line of methods, such as Text2Mesh~\cite{text2mesh}, Tango~\cite{tango}, and CLIP-Mesh~\cite{clipmesh}, utilize CLIP-space similarities as an optimization objective to create novel 3D shapes and textures.
Gao et al.~\cite{get3d} trains a model to generate shape and texture via a DMTet~\cite{dmtet} mesh extractor and 2D adversarial losses.
A recent approach called DreamFusion~\cite{dreamfusion} introduces the use of pre-trained diffusion models to generate 3D NeRF~\cite{nerf} models based on a given text prompt.
The key component in DreamFusion is the score distillation sampling (SDS), which uses a pre-trained 2D diffusion model as a critique to minimize the distribution of the predicted and ground-truth Gaussian noise, thus the 3D scene can be optimized for desired shape and texture generation.

In the context of texture generation, Latent-NeRF~\cite{latentnerf} demonstrated how to employ SDS loss in the latent space of the diffusion model to generate textures for 3D meshes and then decoded to RGB for the final colorization output.
Besides, both TEXTure~\cite{TEXTure} and Text2Tex~\cite{Text2Tex} proposed a non-optimization method with progressive updates from multiple viewpoints to in-paint the texture over the 3D mesh models.

Human-specific shape and texture generation methods also follow the same ideas that use either CLIP similarity between the generated human image and the textural descriptions~\cite{avatarclip} or directly leverage SDS for iterative shape and texture generation~\cite{dreamhuman, avatarbooth, avatarcraft}.
Besides, they also employ human body model prior, \ie, SMPL~\cite{smpl}, for effective human avatar generation.
However, most generated human textures are over-smooth and of low quality, which we argue is caused by the unstable guidance provided by SDS.


\section{Method}

In this section, we start with an overview of SDS.
We then introduce Denoised Score Distillation (DSD), which uses an extra negative pair of image-text to guide gradient direction, thereby generating detailed textures that align with the input text.
Finally, we employ depth signals in the diffusion process for complex surface texturing and employ a geometry-aware rendering function for photorealistic human texture generation.
The overall pipeline is shown in Figure~\ref{fig:painthuman_pipeline}.

\subsection{SDS Overview}
Given an input image $\mathbf{x}$ with a latent code $\mathbf{z}$, a conditioning text embedding $y$, a denoising U-Net $\epsilon_{\phi}$ with model parameters $\phi$, a uniformly sampled timestep $t \sim \mathcal{U}(0, \mathbf{I})$, and a Gaussian noise $\epsilon \sim \mathcal{N}(0, \mathbf{I})$, the diffusion loss is:
\begin{equation}
    \mathcal{L}_{\text{Diff}}(\mathbf{z}, y, t) = w(t)\| \epsilon_{\phi}(\mathbf{z}_{t}, y, t) - \epsilon \|^{2}_{2},
    \label{eq:painthuman_diff_loss}
\end{equation}
where $w(t)$ is a weighting function depending on $t$, and $\mathbf{z}_{t}$ refers to the noisy version of $\mathbf{z}$ via an iterative forward diffusion process given by $\mathbf{z}_{t} = \sqrt{\alpha_{t}} \mathbf{z} + \sqrt{1-\alpha_{t}}\epsilon$, with $\alpha_t$ being the noise scheduler.
For high quality generation, classifier-free guidance (CFG)~\cite{cfg} is used, which jointly learns text-conditioned and unconditioned models via a scale parameter $\omega$. During inference, the two models are used to denoise the image as follows:
\begin{equation}
    \hat{\epsilon}_\phi\left(\mathbf{z}_t, y, t\right)=(1+\omega) \epsilon_\phi\left(\mathbf{z}_t, y, t\right)-\omega \epsilon_\phi\left(\mathbf{z}_t, t\right).
    \label{eq:painthuman_cfg}
\end{equation}
Given a differentiable rendering function $g_{\theta}$, the gradient of diffusion loss with respect to model parameters $\theta$ is:
\begin{equation}
    \nabla_\theta\mathcal{L}_{\text{SDS}} = w(t)\left(\hat{\epsilon}_\phi\left(\mathbf{z}_{t}, y, t\right)-\epsilon\right) \frac{\partial \mathbf{z}_{t}}{\partial \theta},
\label{eq:painthuman_sds_loss}
\end{equation}
where we have omitted the U-Net Jacobian term as shown in \cite{dreamfusion}.
The purpose of SDS is to generate samples via optimization from a text-guided diffusion model.
However, we argue that SDS only presents poor guidance on input text prompt and the generated 2D image, hence, in the following, we propose a new loss design to increase the generation quality.

\subsection{Denoised Score Distillation}
Given a textureless human avatar, our task is to generate surface textures conditioned on input texts.
Due to SDS and neural representation of 3D avatar~\cite{nerf}, zero-shot human texture generation is made possible.
%
We observe that using SDS only for human texturing can cause over-smoothed body parts and cannot be fully semantically aligned to the input text.

We address the issue brought by SDS by proposing a new method, Denoised Score Distillation (DSD), for detailed human avatar texturing of high quality.
Specifically, when presented with input text embedding $y$ and the corresponding image $\mathbf{x}$ with the latent code $\mathbf{z}$, we aim to refine the gradient $\nabla_\theta\mathcal{L}_{\text{SDS}}$ in Eq.~\ref{eq:painthuman_sds_loss} to a direction, so that the rendered avatar contains a detailed texture mapping that is semantically aligned to the input text.
Mathematically, our DSD score function is formulated as:
\begin{equation}
    \mathcal{L}_{\text{DSD}} = w(t)\big(\| \epsilon_{\phi}(\mathbf{z}^{i}_{t}, y, t) - \epsilon \|^{2}_{2} - \lambda\| \epsilon_{\phi}(\hat{\mathbf{z}}^{i-1}_{t}, \hat{y}, t) - \epsilon \|^{2}_{2} \big),
    \label{eq:painthuman_dsd}
\end{equation}
where we introduce a \textit{negative} pair of image with latent code $\hat{\mathbf{z}}$ and text with embedding $\hat{y}$.
$\lambda$ is a weighting parameter. Both $\mathbf{z}^{i}_{t}$ and $\hat{\mathbf{z}}^{i-1}_{t}$ have a superscript $i$ indicating the training iteration and share the same timestep $t$ and noise $\epsilon$, allowing us to use the same U-Net for noise prediction.
Then the gradient of $\mathcal{L}_{\text{DSD}}$ over the model parameter $\theta$ is:
\begin{equation}
    \begin{gathered}
    \nabla_\theta\mathcal{L}_{\text{DSD}} = w(t)\big(\hat{\epsilon}_\phi\left(\mathbf{z}_{t}, y, t\right)-\epsilon - \lambda(\hat{\epsilon}_\phi\left(\hat{\mathbf{z}}_{t}, \hat{y}, t\right)-\epsilon)\big) \frac{\partial \mathbf{z}_{t}}{\partial \theta} \\
     = w(t)\big(\hat{\epsilon}_\phi\left(\mathbf{z}_{t}, y, t\right) - \lambda\hat{\epsilon}_\phi\left(\hat{\mathbf{z}}_{t}, \hat{y}, t\right) - (1-\lambda)\epsilon \big) \frac{\partial \mathbf{z}_{t}}{\partial \theta},
    \end{gathered}
\label{eq:painthuman_dsd_score}
\end{equation}
where we have omitted the U-Net Jacobian matrix following~\cite{dreamfusion}.

As depicted in Figure~\ref{fig:painthuman_pipeline}, we employ the negative image $\hat{\mathbf{x}}^{i-1}$ derived from the preceding training iteration, where we consider $\hat{\mathbf{x}}^{i-1}$ a negative version of $\mathbf{x}^{i}$ as it contains more noise signals.
The inclusion of the negative image within the computation process of $\nabla_\theta\mathcal{L}_{\text{DSD}}$ yields two significant advantages.
Firstly, $\hat{\mathbf{z}}_{t}^{i-1}$ can reinforce the memory of the rendered human image during long time training, so that the final output can still be semantically aligned to the input text.
Secondly, the incorporation of the negative image improves the model's capacity to learn complex geometries, thus facilitating the generation of clear boundaries between varying garment types.
For negative prompts, we use the common prompts such as \textit{disfigured}, \textit{ugly}, etc.
However, we would adapt existing prompts based on a test run, infusing refined negative prompts based on the observed output.
For instance, if artifacts emerge within rendered hand regions, we append ``\textit{bad hands}" to the prompt set.
In contrast to the indirect application of negative prompts in Stable Diffusion, we inject the negative prompt embedding directly into $\nabla_\theta\mathcal{L}_{\text{DSD}}$. This strategy effectively minimizes artifact presence in the rendered human images, thereby enhancing the quality of the generated output.

Through the integration of both negative image and prompts, we successfully manipulate the existing SDS gradient in Eq.~\ref{eq:painthuman_sds_loss} to guide the model convergence towards a mode that yields highly detailed and qualitative textures, which also remain semantically aligned to the input text. Further analyses and insights into this approach are provided in our ablation study.




\subsection{Geometry-aware Texture Generation}
\noindent{\textbf{Geometry Guidance in DSD.}}
To accurately texture the complex garment details, we compute and leverage the corresponding depth map as a fine-grained guidance. Therefore, we employ a pre-trained depth-to-image diffusion model~\cite{sd} rather than the general version, so that the generated avatar could follow the same depth values of the given surface mesh.
As shown in Figure~\ref{fig:painthuman_dsd}~(b), although the rendered human image presents textures that are not semantically aligned to the input text as the belt region is not clearly textured, utilizing the depth-aware diffusion model ensures the generated texture preserve more geometric details and semantically aligned to the given geometry.

\noindent{\textbf{Shading Model for Rendering.}}
Following the idea of physically based rendering (PBR), which models and renders real-world light conditions and material properties, we estimate surface materials by leveraging SV-BRDFs for human image rendering:
\begin{equation}
    R(\mathbf{x}_{p}, \mathbf{l})=\int_{H} L_i(\mathbf{l}) (f_d + f_s) \left(\mathbf{l} \cdot \mathbf{n}\right) \mathrm{d}\mathbf{l},
\label{eq:pbr_rendering}
\end{equation}
where $L_i(\mathbf{l})$ is the incident radiance, and $H=\{\mathbf{l}: \mathbf{l}\cdot \mathbf{n} \geq 0\}$ denotes a hemisphere with the incident light and surface normal $\mathbf{n}$.
$f_s$ and $f_d$ are diffuse and specular SV-BRDFs, respectively.

In particular, we follow~\cite{realshading} to employ simple diffuse model for a low computational cost.
The diffuse SV-BRDF is mathematically expressed as follows:
\begin{equation}
    f_{d}(\mathbf{x}_{p}) = \frac{\mathbf{k}_d}{\pi},
\end{equation}
where $\mathbf{k}_d$ is the diffuse term which can be learned based on 3D vertex positions.
For specular SV-BRDF estimation, we use a microfacet specular shading model as in~\cite{realshading} to characterize the physical properties of the mesh surface:
\begin{equation}
    f_{s}(\mathbf{l}, \mathbf{v}) = \frac{DFG}{4(\mathbf{n} \cdot \mathbf{l})(\mathbf{n} \cdot \mathbf{v})},
\label{eq:specular_brdf}
\end{equation}
where $\mathbf{v}$ is the view direction.
$D$, $F$ and $G$ represent the normal distribution function, the Fresnel term and geometric attenuation, respectively.
We further choose the Disney BRDF Basecolor-Metallic parametrization~\cite{disney} for physically accurate rendering.
Specifically, the specular reflectance term $\mathbf{k}_{s} = m \cdot \mathbf{k}_{d} + (1-m) \cdot 0.04$, where the diffuse term $\mathbf{k}_{d}$, the roughness term $r$ and metallic term $m$ can be estimated via our proposed SV-BRDF network give surface points $\mathbf{x}_{p}$: $\sigma(\gamma(\mathbf{x}_{p})) = [\mathbf{k}_{d}, r, m]$.
$\gamma$ is a coordinate-based network~\cite{ngp} and $\sigma$ is a parameterized SV-BRDF estimation model.
We also utilize a differentiable split-sum approximation for Eq.~\ref{eq:pbr_rendering} to maintain the differentiability in the rendering process.
Moreover, we regularize the material learning process following~\cite{nerfactor}, which results in a smooth albedo map.

\begin{figure*}[t]
    \centering
    \includegraphics[width=0.85\linewidth]{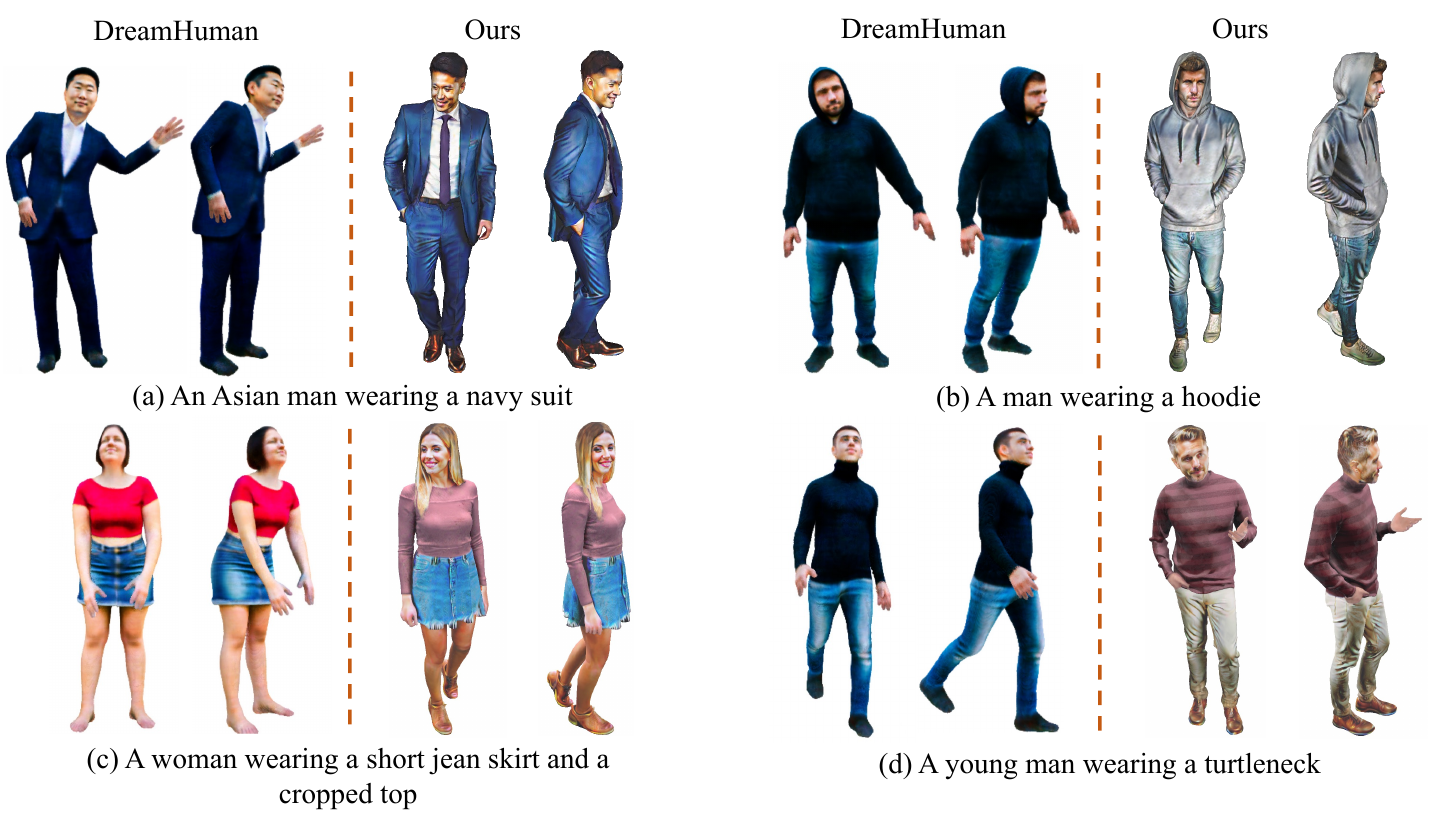}
    \caption{
    Qualitative comparisons with DreamHuman~\cite{dreamhuman}. As DreamHuman is not publicly available, we pick similar mesh models from Renderpeople~\cite{renderpeople} and download the results from the published paper.
    }
    \vspace{-3mm}
    \label{fig:dreamhuman}
\end{figure*}

\noindent{\textbf{Semantic Zoom.}}
Human perception is particularly sensitive to distortions and artifacts in facial features.
However, texturing human avatars in a full-body context often results in degraded facial details. To address this issue, we enhance the human prior during the optimization process by semantically augmenting the prompt~\cite{avatarclip}. For instance, we prepend ``the face of'' to the beginning of the prompt to direct more attention to this region. Simultaneously, every four iterations, we shift the look-at point of the camera to the face center and semantically zoom into the facial region, which refines facial features and improves the overall perception of the rendered avatar.

\section{Experiments}
\subsection{Experimental Setup}
\noindent{\textbf{Baseline Models.}}
We compare our model to recent state-of-the-art baseline models, including Latent-Paint~\cite{latentnerf}, TEXTure~\cite{TEXTure}, and Fantasia3D~\cite{fantasia3d} with the appearance modeling part only. We modify Fantasia3D to ensure the vertex positions remain fixed whiling generating textures.
We also compare our model performance with a recent method for realistic human avatar generation, DreamHuman~\cite{dreamhuman}, to further validate the effectiveness of our design. Although the human mesh model is not publicly available, we use the same text prompts as in DreamHuman to evaluate the quality of human textures with similar human mesh models.

\noindent{\textbf{Implementation Details.}}
We use nvdiffrast~\cite{Laine2020diffrast} to render the depth image from the given human model.
Our SV-BRDF network contains a coordinate-based learning module~\cite{ngp} followed by an MLP with 32 hidden units.
Our model is trained on a single Nvidia RTX 3090 GPU for 10000 iterations to achieve the best performance, where we employ AdamW optimizer with a learning rate of $1 \times 10^{-2}$.
Once the training process is completed, we uniformly sample 100 camera poses for the visualization of rendered human avatars.

\noindent{\textbf{Camera Settings.}}
We define the camera position $(r, \theta, \varphi)$ following Fantasia3D~\cite{fantasia3d} in the spherical coordinate system, where $r=3$, $\theta \in \mathcal{U}(-\pi / 18, \pi /4)$, and $\varphi \in \mathcal{U}(\pi /7, \pi /4)$.
For the additional face camera defined for semantic zooming, different human mesh models have different settings. With the current design, we manually look for the center of the face region and uniformly sample the values for $\theta$ and $\varphi$: $\theta \in \mathcal{U}(-\pi/4, \pi/4)$ and $\varphi \in \mathcal{U}(7\pi/18, 5\pi/9)$.

\begin{figure*}[ht]
    \centering
    \includegraphics[width=0.9\linewidth]{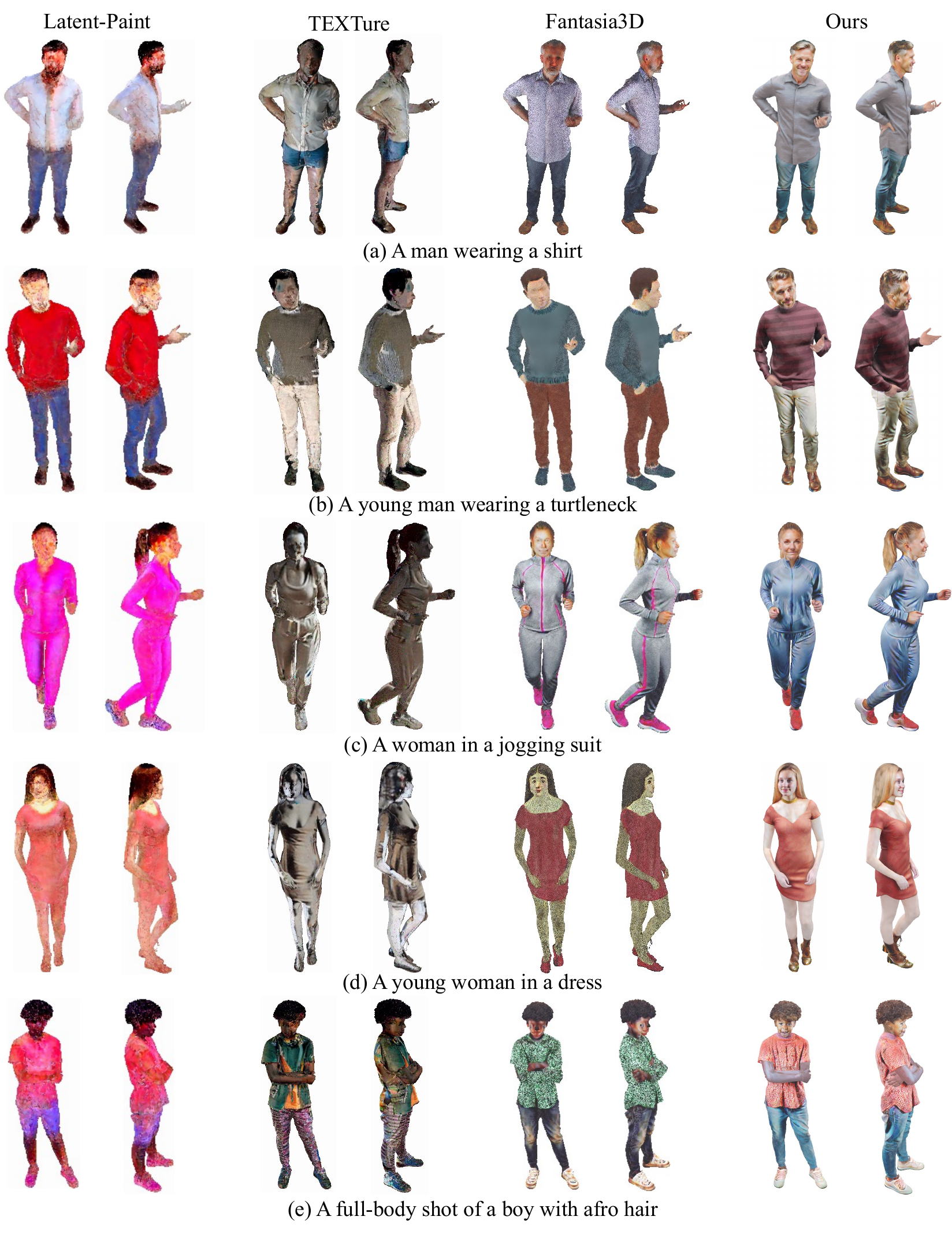}
    \caption{
    Qualitative comparisons on RenderPeople~\cite{renderpeople} for textured human avatars against Latent-Paint~\cite{latentnerf}, TEXTure~\cite{TEXTure}, and Fantasia3D~\cite{fantasia3d}.
    Our generation contains the best texture quality with high-frequency details and consistent with input textual descriptions.
    }
    \label{fig:painthuman_baseline}
\end{figure*}

\begin{table}
    \centering
    \small
    \begin{tabular}{l|c|c} 
    \toprule
    Method       & Mean CLIP Score & $\Delta$ (\%)  \\ 
    \hline
    Latent-Paint  & 24.11           & 19.99\%                      \\
    TEXTure        & 25.34           & 14.17\%                      \\
    Fantasia3D    & 27.10           & 6.75\%                       \\
    PaintHuman (Ours)                            & \textbf{28.93 }          & -                          \\ 
\hline
\hline
DreamHuman  & 25.79           & 12.25\%                       \\
\textbf{PaintHuman (Ours)}                           & \textbf{28.95}           & -                           \\
\bottomrule
\end{tabular}
\caption{Quantitative comparisons of mean CLIP score between baseline models and ours for textured human avatars.
$\Delta$ denotes the percentage by which our model outperforms the indicated method.}
\label{tab:clip_diff}
\vspace{-1mm}
\end{table}

\subsection{Qualitative Analysis}
\begin{figure*}
    \centering
    \includegraphics[width=0.92\linewidth]{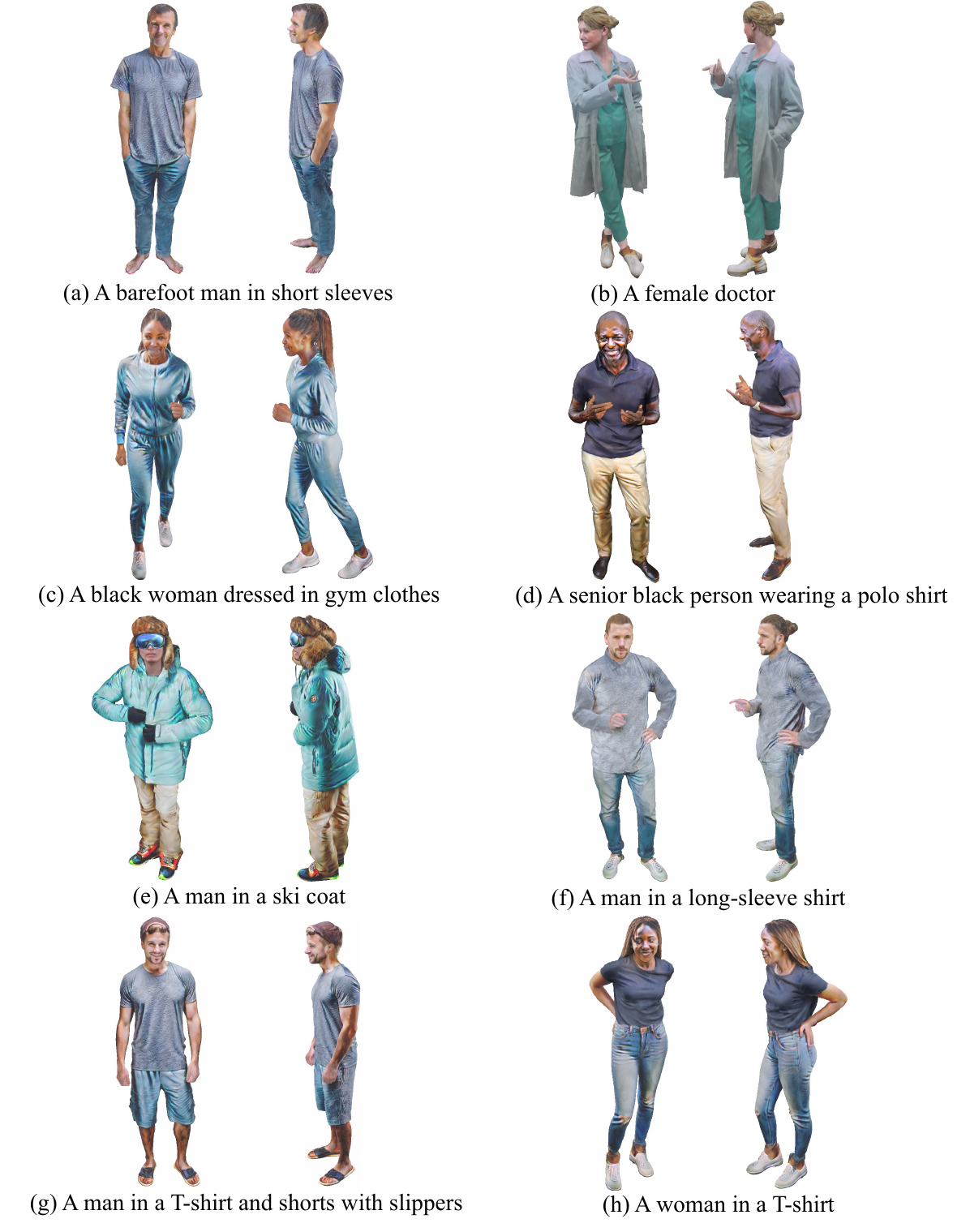}
    \caption{More rendered results on RenderPeople~\cite{renderpeople} for textured human avatars guided by textual descriptions, where the corresponding input text is shown at the bottom of each sample.}
    \label{fig:painthuman_more_results}
\end{figure*}

As depicted in Figure~\ref{fig:painthuman_baseline}, we compare our qualitative results against baseline models.
Latent-Paint is unable to capture the semantics of the objects, which results in failed or blurry textured avatars. 
TEXTure generates relatively better results than Latent-Paint, while it still suffers from inconsistent textures. 
Fantasia3D performs well given certain input texts as in  Figure~\ref{fig:painthuman_baseline}~(a) and (c).
By using SDS which causes the unstable loss gradient direction, Fantasia produces unrealistic samples with noisy textures in most cases.
In contrast, our model can output realistic textured avatars with high-quality and detailed textures, which are aligned to input texts and consistent with the geometry.
We further compare our results with DreamHuman in Figure~\ref{fig:dreamhuman}. We observe that using the same text input, our model generates textured avatars with more high-frequency details, such as the cloth wrinkles, which is different from DreamHuman where the textures are over-smoothed.
Moreover, in both experiments, our model can consistently generate high-quality human faces.
More results are displayed in Figure~\ref{fig:painthuman_more_results}.


\begin{table}
\centering
\small
\begin{tabular}{l|c|c} 
\toprule
Method       & Score & $\Delta$ (\%)  \\ 
\hline
Latent-Paint & 1.21$\pm$0.70           & 148.76\%                      \\
TEXTure      & 1.28$\pm$0.60           & 135.16\%                      \\
Fantasia3D   & 1.76$\pm$0.70           & 71.02\%                       \\
DreamHuman   & 2.83$\pm$0.82           & 6.36\%                       \\
\hline
PaintHuman (Ours)         & \textbf{3.01$\pm$0.95}           & -      \\
\bottomrule
\end{tabular}
\caption{
User study results of baseline models and ours.
$\Delta$ denotes the percentage by which our model outperforms the indicated method.}
\label{tab:userstudy}
\vspace{-4mm}
\end{table}

\subsection{Quantitative Analysis}
To investigate the alignment between the rendered human avatars and the input texts, we leverage the CLIP score~\cite{clip}.
As shown in Table~\ref{tab:clip_diff}, we compare our method against the baseline models and report the mean CLIP score.
Specifically, we generate six frontal images from all textured avatars, each separated by a 30-degree interval.
we observe that our model outperforms all baseline models, where our result is higher than Latent-Paint by the largest margin of around 19.99\%.
Such improvements demonstrate that our proposed DSD is capable of generating more realistic textures on complex human meshes, and is better aligned to the input texts.




\subsection{User Study}
We conduct user study to analyze the quality of the generated textures and the fidelity to the input text. 
Questions are designed from three perspectives: the generation quality of the rendered garments, the generation quality of the human body texture, and the generation quality of the face area.
In the designed questionnaire, we ask each participant to evaluate the given rendered human images from all three perspectives and give their scores in an objective point of view.
Data processing has been done to clean the collected raw data from all the users.
Specifically, we remove the erroneous data where 1 is taken as the most satisfied and 5 the least satisfied. For each question, we remove the highest and the lowest scores and compute the mean and standard deviation values.

We select 4 samples for each baseline method and ask users to rate their overall quality, including texture quality and alignment between text and rendered results, 1 for least satisfied and 5 for most satisfied.
Collected results are reported in Table~\ref{tab:userstudy} including mean scores and standard deviation values, which indicate that our method outperforms the baselines. 


\begin{figure}[t]
    \centering
    \includegraphics[width=0.9\linewidth]{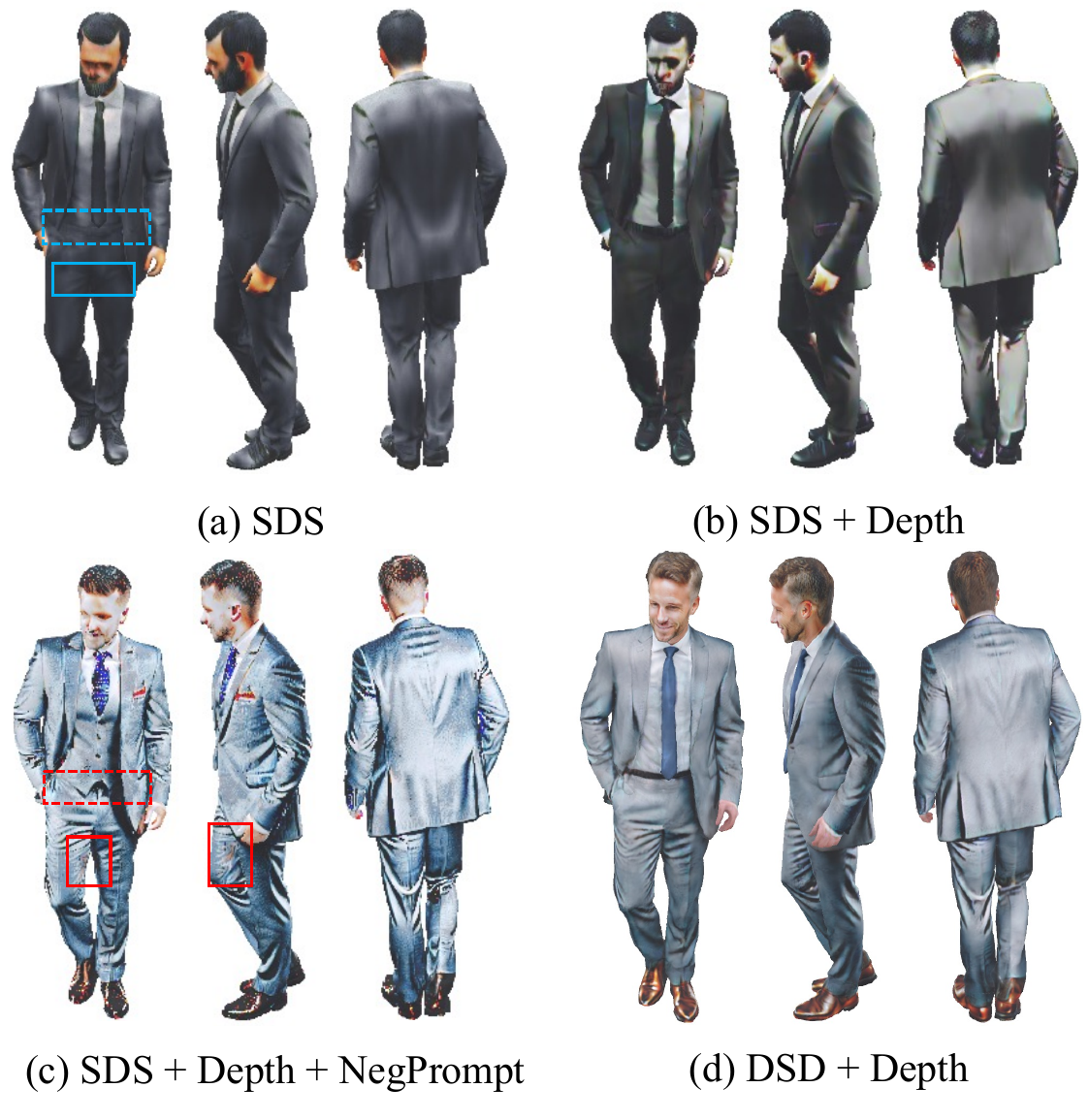}
    \caption{
    Rendered results of textured human avatars based on different settings. 
    } \label{fig:painthuman_dsd}
\end{figure}

\begin{figure}
    \centering
    \includegraphics[width=1\linewidth]{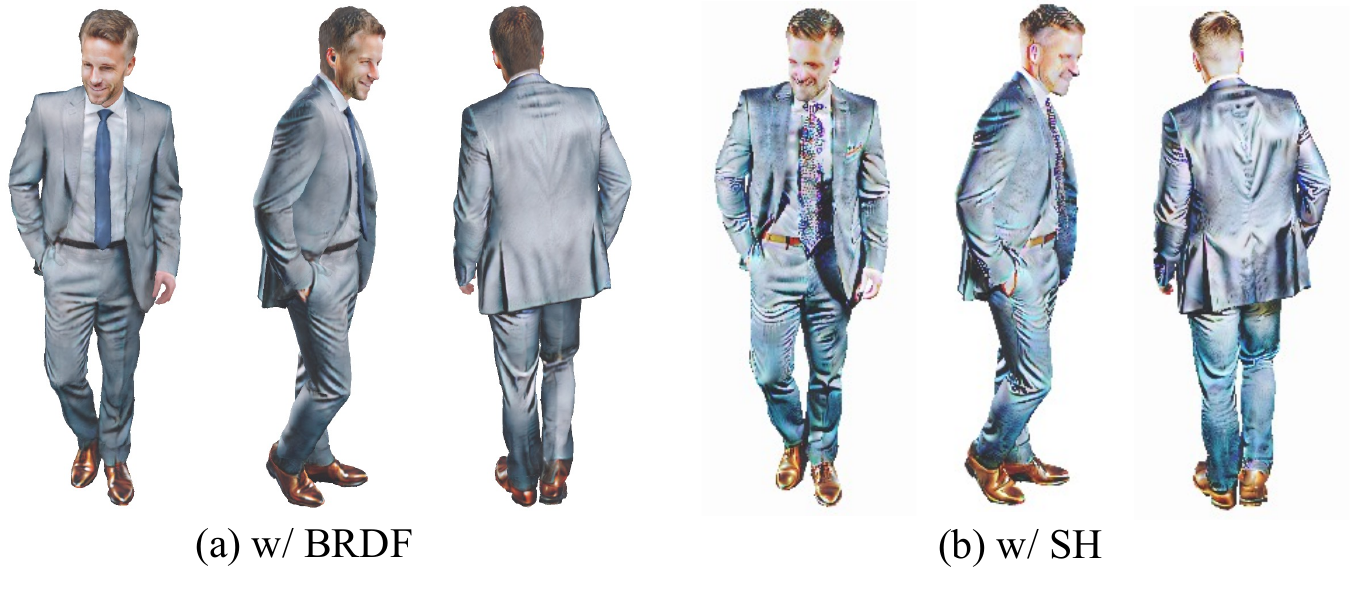}
    \caption{
    Ablation study on different shading models. (a) Fully-proposed method with BRDF; (b) Fully-proposed method with Spherical Harmonic model (SH).
    } \label{fig:painthuman_shading}
\end{figure}

\subsection{Ablation Study} \label{sec:ablation}
\begin{figure}[t]
    \centering
    \includegraphics[width=0.7\linewidth]{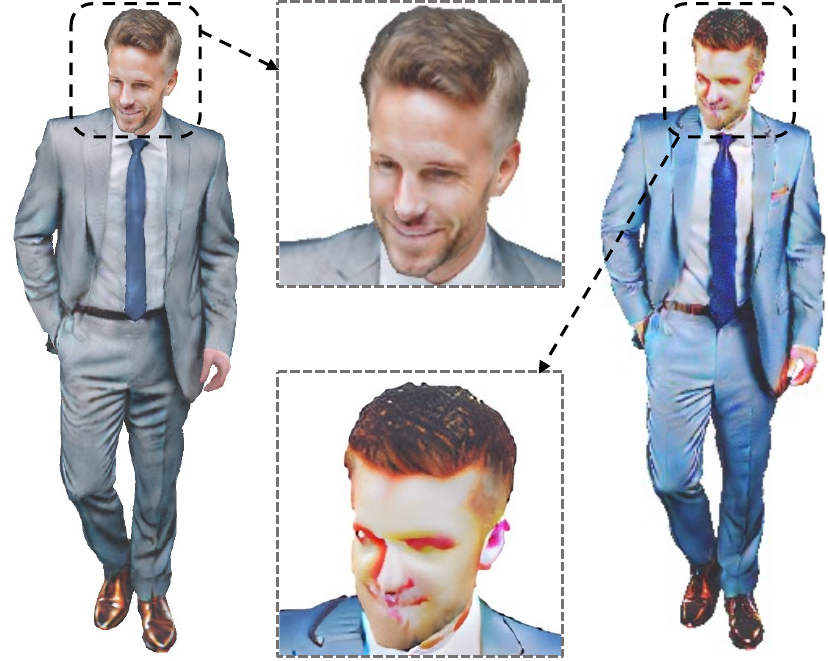}
    \caption{
    Importance of semantic zoom. The left image shows the generated avatar with semantic zoom, while the right image employs no semantic zoom.
    } \label{fig:painthuman_face_semantic}
\end{figure}

To validate the effectiveness of our proposed components, we use ``a man in a suit with a belt and tie" as an example text prompt for the ablation study. Results of alternative settings are shown in Figures~\ref{fig:painthuman_dsd}, \ref{fig:painthuman_shading}, and \ref{fig:painthuman_face_semantic}.

Firstly, the efficacy of our DSD is verified through several comparisons.
As shown in Figure~\ref{fig:painthuman_dsd}(a), we note that employing SDS for human texturing often results in over-smoothed body parts and fails to fully align with the input text semantically, where the belt region is neglected. The addition of depth map guidance in Figure~\ref{fig:painthuman_dsd} also struggles to address this issue.
Moreover, by adding negative prompts, Figure~\ref{fig:painthuman_dsd}(c) demonstrates that the rendered image is able to include more high-frequency details, but is not aligned with the input text, and some parts are devoid of texturing.
In contrast, as shown in Figure~\ref{fig:painthuman_dsd}(d), an image rendered using our DSD effectively mitigates the over-smoothing issue and results in a high-quality, detailed human avatar.

We further examine the effectiveness of BRDF shading model.
As shown in Figure~\ref{fig:painthuman_shading}~(b), we render the result with the Spherical Harmonic model (SH)~\cite{nerd}, resulting in less realistic textures with noticeably noisy color distributions at the borders between different garments. However, using BRDF can give us smooth and clear textures.
%
Finally, as shown in Figure~\ref{fig:painthuman_face_semantic}, our usage of semantic zoom on the face region significantly enhances the overall texture quality. Notably, the method enables the presence of intricate facial features, contributing to a more realistic representation.

\section{Conclusion}
In this work, we introduce PaintHuman, a zero-shot text-to-human texture generation model.
We present a novel score function, Denoised Score Distillation (DSD), which refines gradient direction to generate high-quality, detailed human textures aligned to the input text.
We also leverage geometry signals into DSD for accurate texturing of complex garment details.
To maintain semantic alignment between the mesh and the synthesized texture, we employ a differentiable network to parameterize SV-BRDFs for surface material prediction, which is complemented by physically based rendering for realistic avatar renderings, with facial details refined through semantic zooming.
Our extensive experiments reveal significant improvements in texture generation, validating the effectiveness of our module designs.

\newpage
{\small
\bibliographystyle{ieee_fullname}
\bibliography{ref.bib}
}
\end{document}